\documentclass{article}

\usepackage[preprint]{neurips_2021}

\usepackage[utf8]{inputenc} 
\usepackage[T1]{fontenc}    
\usepackage{hyperref}       
\usepackage{url}            
\usepackage{booktabs}       
\usepackage{amsfonts}       
\usepackage{nicefrac}       
\usepackage{microtype}      
\usepackage{xcolor}         
\usepackage{amsfonts}
\usepackage{subfigure}
\usepackage{graphicx}
\usepackage{booktabs} 
\usepackage[]{breqn}
\usepackage{amsmath}
\usepackage{booktabs}       
\usepackage{lipsum}
\usepackage{textcomp}
\usepackage{adjustbox}
\usepackage{wrapfig}
\usepackage{float}
\usepackage{soul}
\usepackage{caption}
\usepackage{enumitem}
\usepackage{algorithm}
\usepackage{algpseudocode}

\title{Inducing Functions through Reinforcement Learning without Task Specification}

\author{
 Junmo Cho \\
 School of Electrical Engineering, KAIST \\
 Daejeon, Republic of Korea \\
 \texttt{junmokane@kaist.ac.kr} 
 \And
 Dong-Hwan Lee \\
 School of Electrical Engineering, KAIST \\
 Daejeon, Republic of Korea \\
 \texttt{donghwan@kaist.ac.kr} \\
 \AND
 Young-Gyu Yoon \\
 School of Electrical Engineering, KAIST \\
 Daejeon, Republic of Korea \\
 \texttt{ygyoon@kaist.ac.kr} \\
}

\begin{document}

\maketitle


\begin{abstract}

We report a bio-inspired framework for training a neural network through reinforcement learning to induce high level functions within the network. Based on the interpretation that animals have gained their cognitive functions such as object recognition — without ever being specifically trained for — as a result of maximizing their fitness to the environment, we place our agent in an environment where developing certain functions may facilitate decision making. The experimental results show that high level functions, such as image classification and hidden variable estimation, can be naturally and simultaneously induced without any pre-training or specifying them.

\end{abstract}

\section{Introduction}

Advances in reinforcement learning has not only facilitated the development of artificial intelligence for solving challenging problems \citep{Ha2018World, Baida2020Agent, Steven2019Recurrent, Baida2020Never, Mnih2015Dqn}, but it has also bridged the gap between learning in biological systems and learning in artificial systems \citep{Neftci2019Bio, Dabney2020Dopa}. For example, temporal difference learning \citep{Sutton1991Temporal}, —which was established in the machine learning field, — led to the reward prediction error theory of dopamine, which suggests that the phasic activity of dopaminergic neurons encodes the difference between the predicted rewards and the experienced rewards \citep{Schultz1997Neural}. More recently, distributional reinforcement learning \citep{Bellemare2017Dist} led to another interesting hypothesis that suggests that a brain simultaneously predicts future rewards as a probability distribution over possible outcomes; that hypothesis was later supported by animal experiments \citep{Dabney2020Dopa}.
The researchers in \citet{Josh2020neuroetho} implemented a virtual rodent that had access to vision and proprioceptive information, and they used neuroethological modeling to characterize motor activity of the rodent. Such research provides a unique opportunity to access the data that is not available from biological systems, which often leads to valid hypotheses on biological intelligence~\citep{Schultz1997Neural} which in turn contributes to artificial intelligence \citep{Dabney2020Dopa, Banino2018Vector}. 

From the point of view of evolution theory, biological intelligence is just one of the many features that animals have gained through evolution as it provides advantages to the species’ survival and the spread of their genes. In other words, animals and their biological intelligence optimize themselves for survival. For example, animals have attained vision and object recognition capability through evolution as such capabilities have provided a huge survival advantage, but not because they were explicitly trained for such tasks. Similarly, animals can understand that objects continue to exist even with the lack of current sensory clues \citep{Miller2009Object}, and they can predict hidden variables such as the trajectory of an invisible moving target \citep{Barborica2003monkey} —
both of which can facilitate their decision making.

Based on this observation, we conjecture that, through reinforcement learning in an environment where image classification and hidden variable estimation are helpful, such functions will be naturally induced within the network without them ever being specified in the training procedure.
We show that this is indeed possible by training a neural network in a custom environment. 

Specifically, we designed a custom environment
that models an animal's survival task in nature
as a Markov decision process. 
In the environment, the agent encounters discrete events (e.g., a predator or a prey) and receives a corresponding vision input (i.e., an image). 
In addition, there are variables
in the environment that the agent does not have access to.
We find that, through end-to-end
reinforcement learning of an agent with convolutional and recurrent sub-networks, image classification and hidden variable estimation capabilities can be induced within each sub-network. The two capabilities are formally defined as follows:
\begin{itemize}[leftmargin=0.3cm]
\item \textbf{Image classification}: to map the input images to the feature vectors that are linearly separable by their classes.
\item \textbf{Hidden variable estimation}: to yield output values that are linearly proportional to the hidden variables.
\end{itemize}

This suggests that high level functions can be induced in an artificial intelligence system through reinforcement learning without task specification, which opens up the possibility of implementing networks with various cognitive functions.

\section{Related work}
\label{related}

\paragraph{Inducing functions within networks}
Training a neural network can be considered as a procedure of learning a composite function $f(x) = f_{n} \circ f_{n-1} \circ \cdots \circ f_{1}(x)$ to solve a task that is specified through a loss function.
The function learned by a part of the network (i.e., $f_{k}(x)$)
is determined jointly by the loss function,   constraints on the function space of each  $f_{j}, j \in \{ 1,2, \cdots n \}$ and  constraints on the vector space that each $ f_{j}$ projects onto.
In \citet{Krizhevsky2012}, 
the weights of the first convolution layer converged to certain shapes which corresponded to image processing functions (e.g., low-pass or high pass filtering).
In \citet{Kingma2014, Chen2016}, the authors demonstrated that certain parts of the network can learn to disentangle different types of information (e.g., style and content) in the input with guidance through the loss function.
In \citet{Makhzani2016}, an adversarial autoencoder was trained in a semi-supervised manner in which MNIST \citep{deng2012mnist} images were projected to a latent space where the images were clustered by their corresponding digits.
However, inducing high level functions without task specification in a reinforcement learning setting has not been demonstrated.

\paragraph{Bio-inspired reinforcement learning}

Owing to the deep-rooted connection between the reinforcement learning in artificial and biological agents, there has been a number of previous works on bio-inspired reinforcement learning \citep{Neftci2019Bio}.
In ~\citet{Singh2005, Lewis2010Reward}, an intrinsic reward for reinforcement learning was proposed to accelerate learning by setting the agent to receive internal rewards from a critic which was part of the agent. 
\citet{Niekem2010Gen, Niekum2011Evol} also employed the notion of intrinsic reward and used a genetic algorithm to search for alternate reward functions.
\citet{Bellemare2017Dist} designed an agent that represented the future rewards as a probability distribution over multiple possible outcomes.
In \citet{Pontes2019Concept}, 
evolution was modeled as the inheritance and change of network topology of spiking neural networks. 
\citet{Abrantes2020Mimick} proposed a learning method that mimics evolution in which the reward function was slowly aligned with the fitness function.
The idea that penetrates these works is that an advanced form of artificial intelligence may be developed by mimicking certain aspects of biological intelligence.
Our work shares that spirit, but it differs from the previous works in that 
our goal is to induce high level functions 
by placing an agent in an environment
where certain high level functions can help decision making
rather than directly mimicking the evolutionary procedure for training networks.

\paragraph{Reinforcement learning with recurrent networks}

Deep recurrent Q-learning~\citep{Matthew2015drqn} and a deep recurrent policy gradient~\citep{heess2015memory}, which employed an agent with convolutional layers and an LSTM, demonstrated their capability to integrate the information across multiple frames.
Since those studies, recurrent units have been widely employed for reinforcement learning. 
For instance, \citet{Steven2019Recurrent} further improved this method for memory-critical problems by devising two strategies for initialization of the state of recurrent units, rather than naive zero initialization \citep{Matthew2015drqn}.
\citep{Ha2018World} proposed a model-based reinforcement learning method, called the world model, that used a recurrent unit which was motivated by
the fact that a human uses a mental model of the world to predict the future.
The recurrent unit was separately trained in a supervised manner to predict the next output from the convolution layers which was then fed to the following fully connected layer for choosing the action.
Simulated policy learning~\citep{Kaiser2020Atari}
demonstrated excellent performance in a low data regime by alternatively updating the world model and the policy. 
In these methods, the role of the recurrent units was to integrate the information from the vision input across time as opposed to predicting the information outside the observation.

\section{Survival environment}

\begin{figure*}[t]
	\renewcommand{\thesubfigure}{}
	\centering
    \subfigure{\includegraphics[width=0.48\textheight]{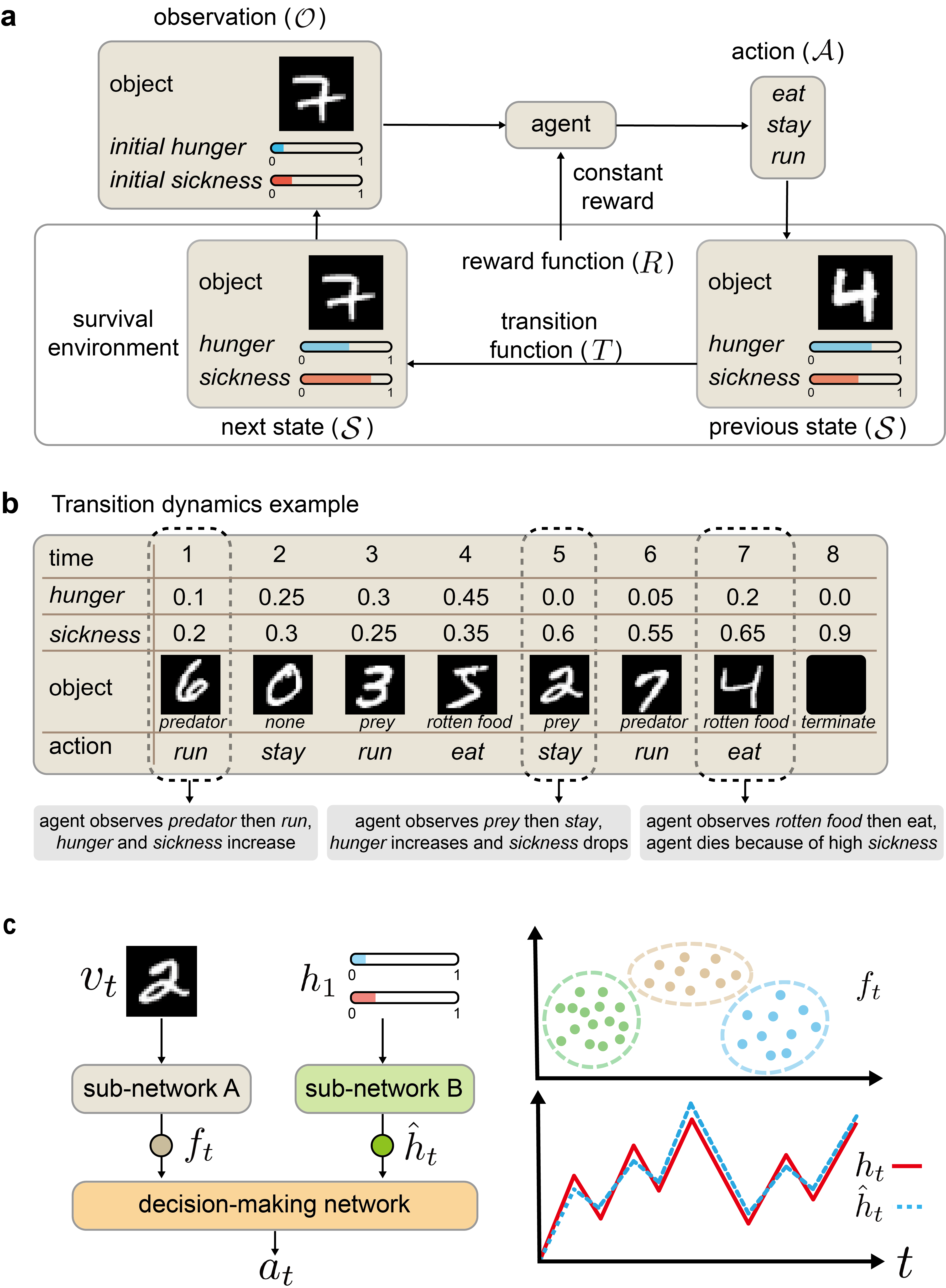}}\\
\caption{
Overview of the survival environment. 
(a) An animal's survival task in nature is modeled as a partially observable Markov decision process. 
At each time point, the agent encounters an object (\textit{none}, \textit{predator}, \textit{prey} or \textit{rotten food}) represented as an image.
In addition, there are two hidden variables (\textit{hunger} and \textit{sickness}) that are not observed by the agent.
The agent should choose an action (\textit{run}, \textit{stay} or \textit{eat}) to maximize its life span.
The image and the initial value of the hidden variables are given to the agent.
(b) A transition dynamics example. Each action has an opportunity cost
and hence, it is not possible to minimize the risks from all sources at the same time.
(c) The agent receives the image and the initial value of the hidden variables (left).
The goal is to induce image classification and hidden variable estimation capabilities within the network
without any pre-training or specification (right).
} \label{fig1}
\vspace{-7pt}
\end{figure*}

\subsection{Formal description of the environment}

We built a custom environment in which an agent makes a sequence of decisions to maximize the length of its life span.
The environment is modeled as a partially observable Markov decision process with a tuple $(\mathcal{S}, \mathcal{A}, T, R, \Omega, \mathcal{O})$ (Figure~\ref{fig1}(a)),
where the state space $\mathcal{S}$ is constructed as a tuple of a vision input, denoted by $v_t$, (i.e., image at time $t$) and additional state variables, denoted by $h_t$, i.e., the state at time $t$ is $(v_t,h_t):= s_t \in \mathcal{S}$. The image corresponds to the object class that the agent encounters, which is one of \textit{none}, \textit{predator}, \textit{prey} and \textit{rotten food}. For each class, we allocated two random digits between $0$ and $9$ (Appendix~\ref{digit_permutation}) and randomly picked 1,000 images for each number from MNIST dataset.
The state variables are \textit{hunger} $H_t$ and \textit{sickness} $S_t$ (i.e., $h_t=(H_t, S_t) \in [0, 1]^{2}$), which depend on their previous values, vision input and action (i.e., $h_t=g(h_{t-1},v_{t-1},a_{t-1})$, where $g$ is the function that determines state variable transition).
The initial values of the state variables are randomly set at the beginning of each episode.
At each time point, the agent chooses one of the following actions: \textit{stay}, \textit{run}, and \textit{eat}, i.e., ${\cal A} = \{\rm \textit{stay}, \textit{run}, \textit{eat}\}$. A constant reward $r_t=1$ is given at each time step $t$ to set the goal of the agent as simply making the episodes as long as possible.

Following the observation function $\mathcal{O}$, only partial information is given to the agent which are the vision input $v_t = \mathcal{O}(v_t,h_t)$ except for time $t=1$, where $(v_1,h_1) = \mathcal{O}(v_1,h_1)$, and the state variables $h_t$ are hidden. For this reason, $h_t$ will be called the hidden variables throughout the paper. 
We note that this is different from most previous reinforcement learning settings with convolutional and recurrent units \citep{Ha2018World, Matthew2015drqn, Steven2019Recurrent, Baida2020Never} in which the recurrent units are used for the prediction of the future vision inputs. 
The complete information on the environment including the transition function $\mathcal{T}$ is given in
Appendix~\ref{env_detail}.

\subsection{Biological interpretation of the environment}

Our custom environment can be interpreted as follows. The goal of the agent — which models an animal — is to survive as long as possible, where the possible causes of death are being eaten by a \textit{predator} and a high \textit{hunger} or \textit{sickness} level. Each action is designed to have an opportunity cost (e.g., \textit{running} decreases the chance to be eaten by a \textit{predator} but increases \textit{sickness}, \textit{eating} \textit{rotten food} decreases \textit{hunger} but increases \textit{sickness}) as it would in nature (Figure~\ref{fig1}(b)), and therefore it is not possible to minimize the risk from all sources at the same time. Consequently, the optimal action at each state depends not only on the object it encounters, but also on the hidden variables (e.g., for a \textit{predator} encounter, choosing \textit{stay} over \textit{run} may be better when \textit{sickness} is high). 
Importantly, the object class information is indirectly given to the agent as an image and the hidden variables are not accessible to the agent. This is similar to the situation where an animal in nature (a) has to process sensory input (e.g., visual, auditory, and olfactory signals) to determine what it is encountering and (b) has to predict unavailable information  to make the decisions using its own memory. 
Essentially, we are placing an agent in an environment
where certain high level functions (image classification and hidden variable estimation) can help the decision making
to induce the functions (Figure~\ref{fig1}(c)).

\section{Survival agent}

\subsection{Agent model}
\label{agent}

\begin{figure*}[t]
	\renewcommand{\thesubfigure}{}
	\centering
    \subfigure{\includegraphics[width=0.61\textheight]{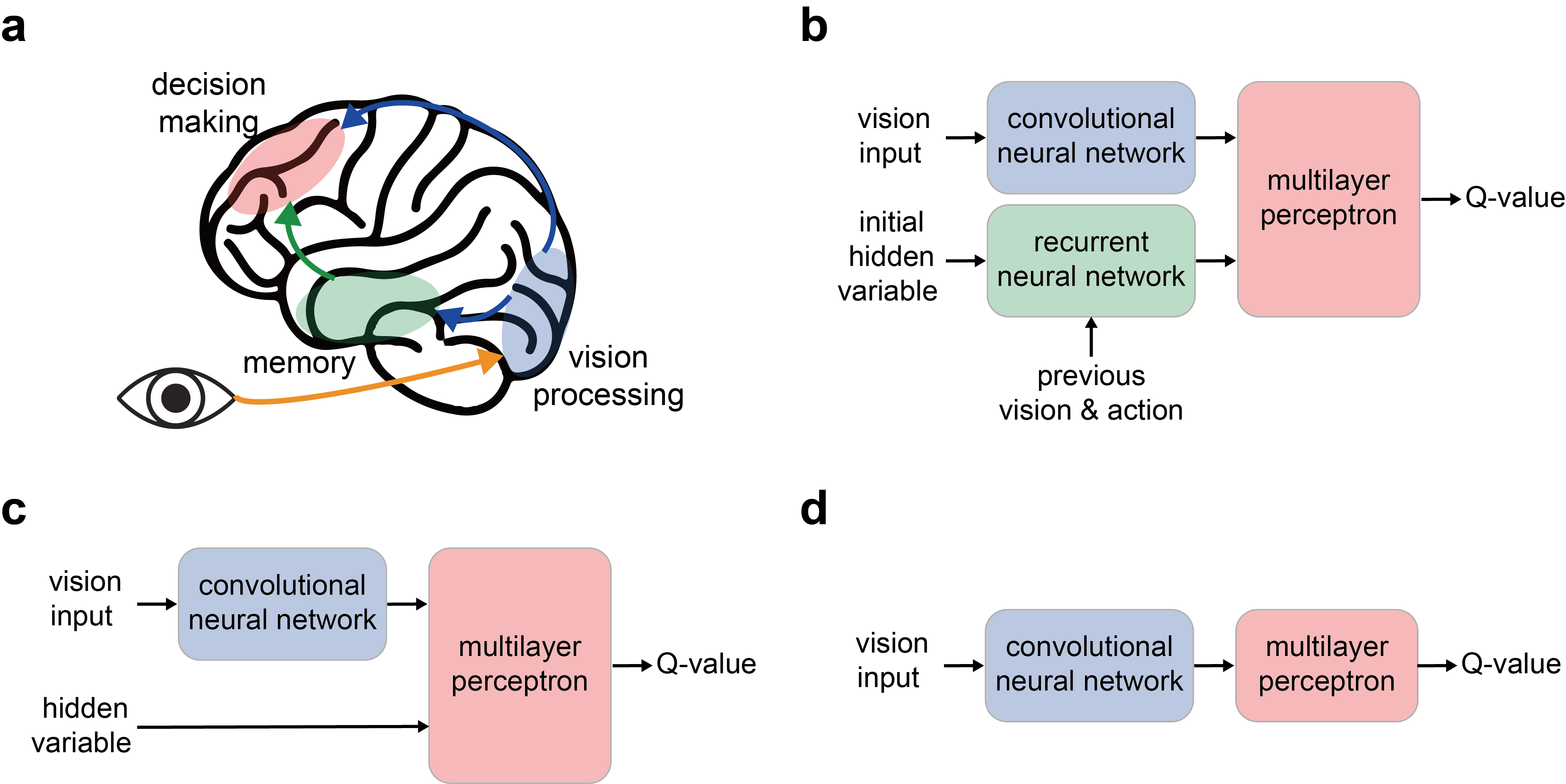}}\\
\caption{(a) Abstract illustration of biological intelligence. 
Visual and memory information is relayed to 
the cerebral cortex to make the decisions.
(b) The network architecture of our agent M1. 
A convolutional neural network (CNN) receives vision inputs from the environment
and a recurrent neural network receives (RNN) the initial value of hidden variables, the previous action and the previous feature vector from the CNN.
The outputs from the CNN and RNN are fed to a multilayer perceptron (MLP) that generates the Q-value of each action.
(c) The network architecture of a baseline model B1.
True hidden variables are directly fed to the MLP. Otherwise the same as M1.
(d) The network architecture of a baseline model B2.
The MLP takes the output only from the CNN. Otherwise the same as M1.
} \label{fig2}
\vspace{-7pt}
\end{figure*}

Our agent, inspired by our cognitive system (Figure~\ref{fig2}(a)), has three components as shown in Figure~\ref{fig2}(b). The first is a convolutional neural network (CNN) that encodes the input image into a 4-dimensional vector. The second is a recurrent neural network (RNN) that makes predictions on hidden variables based on historical information. Lastly, the agent has a multilayer perceptron (MLP) that calculates the Q-value of each action based on the outputs from the CNN and the RNN. 

The network operates as follows.
The CNN takes the image $v_t$ from the environment and generate the output $f_t=\text{CNN}(v_t)$.
When $t=1$, the RNN is not utilized: $h_1$ bypasses the RNN and directly goes to the MLP.
For $t>1$, the RNN takes its previous output, previous feature vector and previous action tuple, $(\hat{h}_{t-1},\text{CNN}(v_{t-1}),a_{t-1})$, and generates the output $\hat{h}_{t}$
whose dimension is the same as the number of hidden variables $(i.e., \hat{h}_{t}=\text{RNN}(\hat{h}_{t-1},\text{CNN}(v_{t-1}),a_{t-1}) \in \mathbb{R}^2)$. The estimated hidden variable $\hat h_t$ and the CNN output $f_t$ are then fed to the MLP to calculate the Q-value
(i.e., $Q(v_t, h_1, a) = \text{MLP}(\text{CNN}(v_t),\hat h_t,a)$).
The details on the network design are in Appendix~\ref{networkdetail}.

\subsection{Baseline models}

In addition to our agent described in Section~\ref{agent}, which we denote by M1, we implemented two baseline models for comparison.
The first baseline model B1 differs from M1 in that the true \textit{hunger} and \textit{sickness} variables are directly fed to the MLP (Figure~\ref{fig2}(c)), and therefore B1 does not have to perform hidden variable prediction. In the second baseline model B2, only the output from the CNN is fed to the MLP without any direct or indirect exploitation of the \textit{hunger} and \textit{sickness} variables (Figure~\ref{fig2}(d)). The architectures of CNNs and the MLPs in M1, B1 and B2 are the same. Therefore, the achievable performance gap between B1 and B2 represents the
importance of the \textit{hunger} and \textit{sickness} variables for choosing the action. Since our agent M1 has an RNN which takes the initial values of the hidden variables, its achievable performance is upper-bounded by that of B1 which has direct access to the variables and lower-bounded by that of B2 which does not have any access to the variables. 

\subsection{Update method}
\label{updatemethod}

We developed a sequential update method that is built upon \citet{Matthew2015drqn} and \citet{Steven2019Recurrent} to train a network with recurrent units with hidden states, which is especially important in our setting as the hidden variables depend heavily on their previous values. 
We first split each episode $e=((v_1,h_1,a_1,r_1),(v_2,a_2,r_2),\ldots,(v_T,a_T,r_T))$ into sub-episodes with a fixed length $L_{seq}$ to construct mini-batchs, where $T$ is the terminal time.
For the sub-episodes that begin at $t \neq 1$, the initial value of the hidden variables was set as the RNN output $\hat{h}_{t-1}$ because the true "initial" values of the hidden states are unknown.
Then, the parameter update is performed by sequentially going through all transactions in the randomly sampled sub-episodes (Appendix \ref{appen_algorithm}). 



\section{Results}
\label{results}

In this section, we present our experimental results, which confirm that high level functions can be induced without task specification, and we compare the results from different settings. The baseline models B1 and B2 were trained with deep Q-learning \citep{Mnih2015Dqn}, and our agent M1 was trained with a variant of deep recurrent Q-learning \citep{Matthew2015drqn, Steven2019Recurrent}, which was described in Section.~\ref{updatemethod}. 
The networks were trained using an Adam optimizer with the learning rate of $3 \times 10^{-4}$ for 1.2 million gradient updates.
The target network was updated with Polyak averaging.
For training B1 and B2, we used replay buffers with a size of $10^{6}$. For M1, 
$10^{5}$ sub-episodes that contain roughly $6 \times 10^{6}$ transactions were saved in the replay buffer.
The networks were implemented using Pytorch and trained on a workstation with two Intel Xeon Scalable Silver 4214R CPUs, four NVIDIA GeForce RTX 2080 Ti GPUs, and 128 GB of RAM. The full implementation of the network and the environment 
is available at \url{https://github.com/anonymous}\footnote{anonymized for review}.


\subsection{Learning curve}
\begin{figure*}[t]
	\renewcommand{\thesubfigure}{}
    \subfigure{\includegraphics[width=\textwidth]{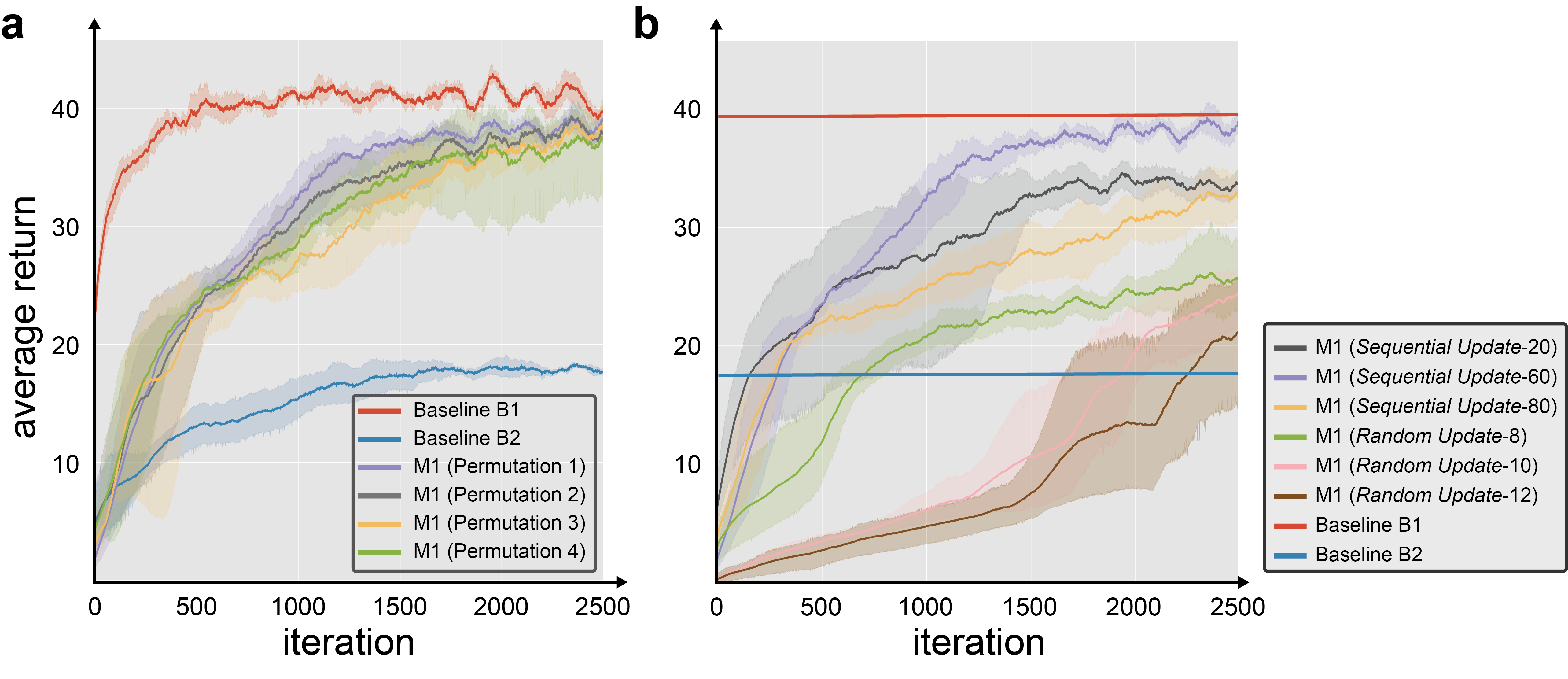}}\\
\caption{(a) Average learning curves of the agents M1, B1 and B2 from five rollouts for each model. B1 with direct access to the hidden variables shows the highest performance (red). 
The average return of M1, which takes the initial value of the hidden variables as the input to its RNN, approaches that of B1 regardless of the digit assignment to each class (purple, grey, yellow, green).
The average return of B2, which does not receive any information on the hidden variables, shows the lowest performance (blue). One iteration corresponds to 500 gradient updates.
(b) Average learning curves of the agents M1 obtained with multiple learning methods and sub-episode lengths.
sequential update (grey, purple, yellow) achieved higher performance than random update (green, pink, brown) proposed in \citet{Steven2019Recurrent}. The performance of B1 (red) and B2  (blue) after training is overlaid for comparison.
}
\label{fig3}
\vspace{-7pt}
\end{figure*}

First we verified the achievable performance of the agent M1 and the baseline models B1 and B2.
Figure~\ref{fig3}(a) shows the average learning curves from five rollouts for each model.
B1 achieved an average return of around 40, whereas B2 had an average return of less than 20. This performance gap between B1 and B2 represents
the importance of taking the hidden variables into account for choosing actions. 
Only for M1, we used four different permutations for the digit assignment to each class (Appendix~\ref{digit_permutation}).
By the end of the training, the average return of M1 was close to that of B1 regardless of the digit assignment
which implies that the images were classified by their class and the RNN was successfully trained and played an important role in the decision making.
In Figure~\ref{fig3}(b), the average learning curves of M1
obtained with different learning methods are compared. 
We evaluated our sequential update method with multiple sub-episode lengths  $L_{seq}=20,60,80$ and the update method proposed in \citet{Steven2019Recurrent} with multiple sequence lengths $L_{ran}=8,10,12$.
Sequential update methods with a sub-episode length of 60 achieved the highest performance.


\subsection{Learning to classify images}
\label{learn_class}

\begin{table}[H]
\vspace{-13pt}
  \caption{
  Classification accuracy of linear support vector machine classifiers. 
  Each classifier is trained to classify  the corresponding object class of the feature vectors from each CNN.
  The classification accuracy embodies the capability of each CNN to map the images to the linearly separable feature vectors.
  CNN-M1 shows consistent accuracy 
  regardless of the digit assignment to each class.
  The average accuracy was obtained from five rollouts with random seeds.}
  \centering
  \vspace{10pt}
  \begin{tabular}{ccc}
    \toprule
    Network & Training accuracy & Test accuracy \\
    \midrule
    CNN-B1 & 99.28 & 95.26      \\
    CNN-B2 & 80.06 & 75.60  \\
    CNN-M1 (permutation 1) & 98.01 & 93.86     \\
    CNN-M1 (permutation 2)  & 97.90 & 93.03 \\
    CNN-M1 (permutation 3)  & 98.20 & 93.01 \\
    CNN-M1 (permutation 4)  & 98.67 & 94.37 \\
    \bottomrule
  \end{tabular}
  \label{table1}
  \vspace{-4pt}
\end{table}

\begin{figure}[!htb]
\centering
\includegraphics[width=0.45\textheight]{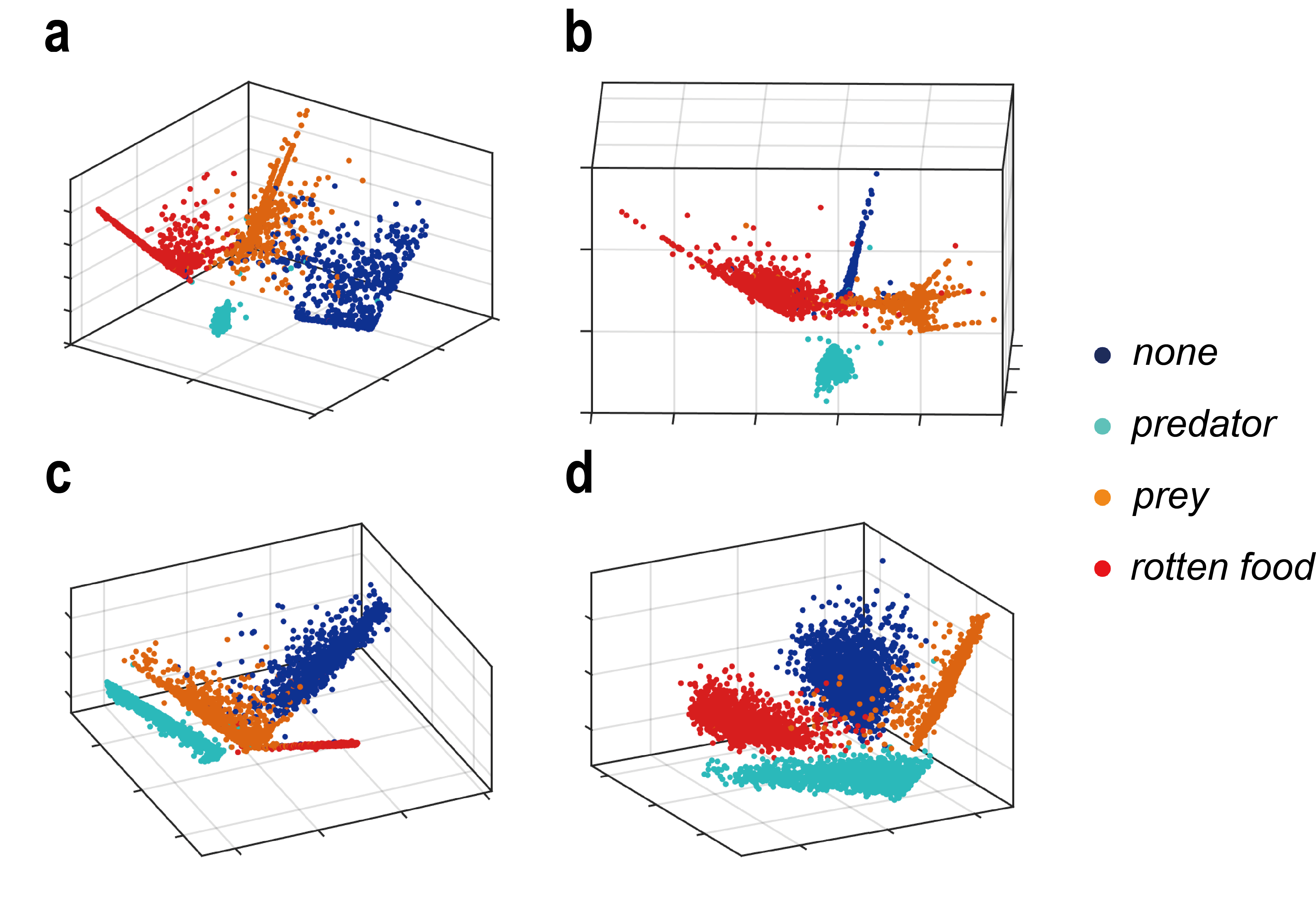}
\caption{Principal component analysis 
is applied to the feature vectors from the CNN to reduce their dimension from 4 to 3 for visualization.
The vectors are linearly separable by their corresponding classes.
(a) Result from CNN-M1 (permutation 1).
(b) Result from CNN-M1 (permutation 3).
(c) Result from CNN-M1 (permutation 4).
(d) Result from CNN-B1.}
\label{fig4}
\end{figure}

To verify whether the network maps the input images to the feature vectors that are linearly separable by their corresponding classes, we extracted the CNNs from M1, B1 and B2, denoted by CNN-M1, CNN-B1 and CNN-B2, respectively, after the training was completed. Then, we fed MNIST images — both that were used and not used for the reinforcement learning — to CNN-M1, CNN-B1 and CNN-B2 and obtained 4-dimensional output vectors. 
Using the output vectors from each network $f_t=\text{CNN}(v_t) \in \mathbb{R}^{4}$ as the input, we trained a linear support vector machine classifier and measured the classification accuracy. 
A total of 8,000 and 32,000 images were used for measuring training accuracy and test accuracy, respectively.

The convolutional network from our agent CNN-M1 achieved an average training and test accuracy of 98.20\% and 93.57\%, respectively,
which shows that the CNN is capable of classifying not only the images used for training, but also the unseen images.
The classification accuracy was consistent across different permutations.
The accuracy from CNN-B1 was slightly higher than that from CNN-M1, whereas the accuracy from CNN-B2 was significantly lower. 
The low performance of CNN-B2 is likely because B2 was not able to comprehend the environment.
The results from all the networks are summarized in Table \ref{table1}.

For visualization of the capability of CNN-M1 and CNN-B1, we applied principal component analysis to the output vectors and reduced their dimension from 4 to 3 (Figure~\ref{fig4}). 
The result shows that the mapped feature vectors are linearly separable by their classes. 
It should be noted that the images with two randomly picked digits from the MNIST dataset were assigned to each class, which indicates that the network learned to cluster images based on their contextual meanings as opposed to
simply clustering visually similar images.
In our setting, the image class and the optimal action are decoupled, (Appendix~\ref{va-decouple})
which also confirms that the CNNs learned to classify images.

\subsection{Learning to predict hidden variables}

\begin{figure*}[!htb]
\vspace{0pt}
	\renewcommand{\thesubfigure}{}
	\centering
    \subfigure{\includegraphics[width=0.6\textheight]{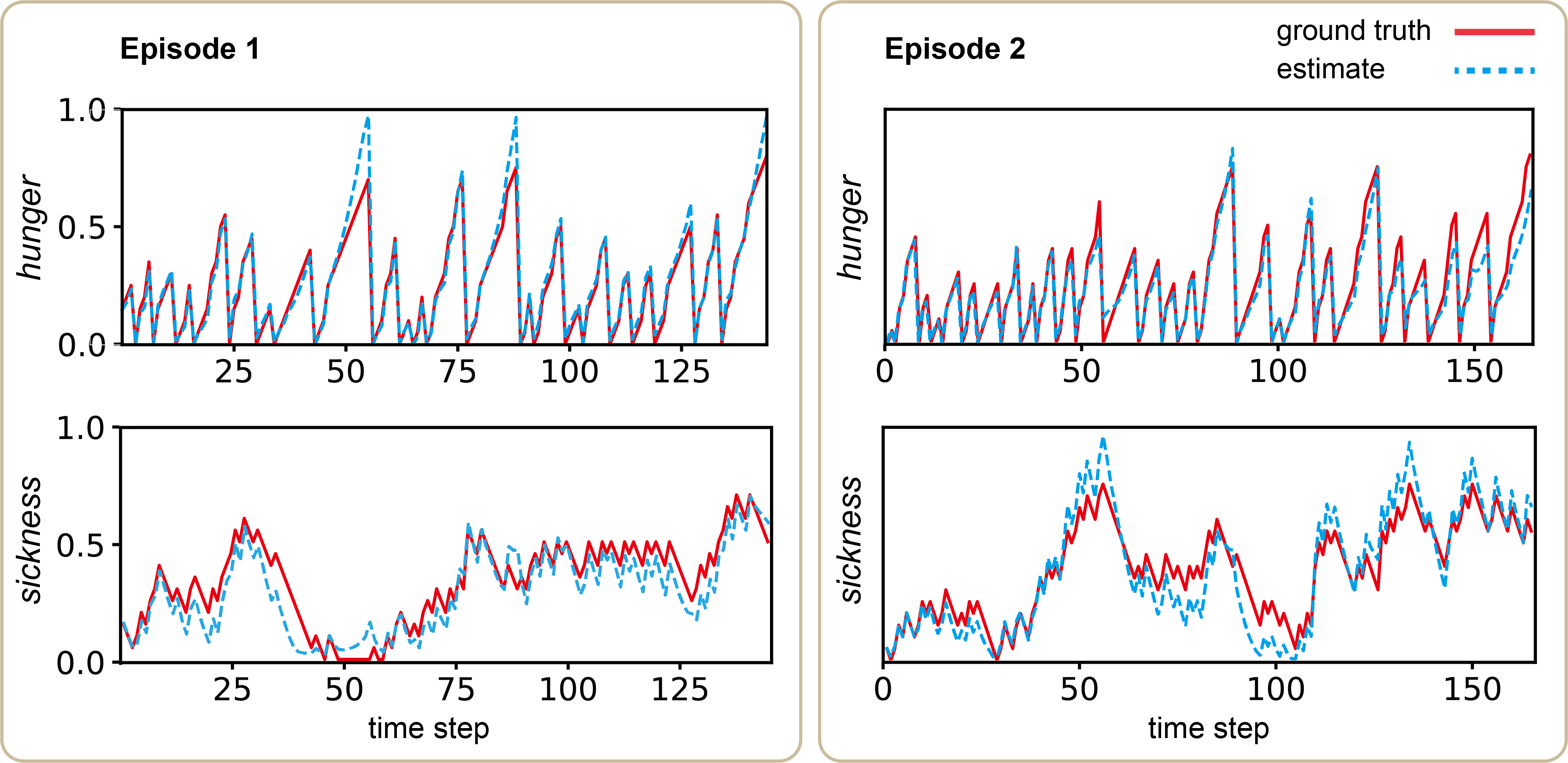}}\\
\caption{Ground truth values of the hidden variables (solid red) and the RNN output (dotted blue).
The time courses of the variables from two episodes are shown.
The average Pearson correlation coefficients between the ground true hidden variables and the RNN output from 200 rollouts were 0.9772 and 0.9450, for \textit{hunger} and \textit{sickness}, respectively.
} \label{fig5}
\end{figure*}

Next, we verified whether the network could yield output values that were linearly proportional to the hidden variables 
using M1 (permutation 1).
After the training was completed,
we collected the test episodes from 200 rollouts.
For each episode, the output from the RNN was  compared with the true hidden variables as shown in Figure~\ref{fig5}.
It is clear that the RNN is capable of predicting the hidden variables despite the fact that it was never specified to do so and it did not have access to the ground truth values of the hidden variables:
We simply let the MLP take the output from RNN as its input. The average Pearson correlation coefficients between the ground true hidden variables and the RNN output
were 0.9772 and 0.9450, for \textit{hunger} and \textit{sickness}, respectively.

\section{Discussion}
\label{discussion}

The main idea of this work is that even high level functions such as image classification can be naturally induced within the network via reinforcement learning.
This is not surprising considering that biological intelligence has developed various cognitive functions due to evolutionary pressure.
Our results indicate that various functions can be induced within artificial networks via reinforcement learning if the environment provides pressure to do so and the network architecture can support the functions.

While this may provide a new framework to develop an artificial intelligence with various cognitive functions, there still are some limitations that need to be overcome.
First, an environment that is as complex as the real world would be necessary to develop an agent that truly mimics biological intelligence. Considering the infeasibility of developing of such environment, it would be necessary to devise a way to let the agent directly interact with the real world.
Second, without knowing what functions the agent may develop beforehand, it is difficult to assess the suitability of a network architecture. 
Potentially, a solution may arise from the field of connectomics~\citep{Li2020} which studies the neural network architectures of biological brains.

It should be noted that unexpected or even unwanted function may be easily developed with this framework due to the absence of task specification. Hence, extra caution is required when  using the induced function for safety or privacy related tasks.


\section{Conclusion}
In this paper, we proposed a bio-inspired framework for training a neural network via reinforcement learning 
to induce high level functions within the network without task specification.
The network learned to classify images and estimate hidden variables simply by placing the agent in an environment where such functions were helpful for making the decisions.
We argue that this closely resembles why and how biological intelligence has developed various cognitive functions and hence, our strategy can be employed and extended to develop artificial intelligence that is truly similar to biological intelligence.


\bibliographystyle{neurips_2021}
\bibliography{neurips_2021}

\appendix
\newpage

\section{Digit assignment to each class}
\label{digit_permutation}

\begin{table}[H]
\vspace{-13pt}
  \caption{
  Digit assignment to each class in each permutation.}
  \centering
  \vspace{10pt}
  \begin{tabular}{ccccc}
    \toprule
     & \textit{none} & \textit{predator} & \textit{prey} & \textit{rotten food} \\
    \midrule
    permutation 1  & 0, 1 & 6, 7 & 2, 3 & 4, 5\\
    permutation 2  & 7, 8 & 5, 3 & 6, 1 & 9, 2\\
    permutation 3  & 2, 5 & 6, 4 & 9, 7 & 8, 3\\
    permutation 4  & 0, 6 & 8, 1 & 3, 7 & 2, 4 \\
    \bottomrule
  \end{tabular}
  \vspace{5pt}
  \label{tabledigitassign}
  \vspace{-7pt}
\end{table}

\section{Environment details} \label{env_detail}

\subsection{Initialization}
When an episode starts, an object ({\textit{none}, \textit{predator}, \textit{prey}}, \textit{rotten food}) is randomly selected with equal probability.
The initial values of \textit{hunger} and \textit{sickness} are randomly drawn from a probability mass function $P(X=0.05k)=\begin{cases}
\frac{1}{11}, \textrm{ if } k \in \{ 0, 1, 2, \cdots, 10\} \\
0, \textrm{otherwise}
\end{cases}$.

\subsection{Object transition table}

\begin{table}[H]
\vspace{-13pt}
  \caption{Object transition table of our custom environment. $a=0.65$ and $b=0.35$ are used. 4-tuple in each cell is the transition probabilities to (\textit{none}, \textit{predator}, \textit{prey}, \textit{rotten food}) for each state-action pair.}
  \centering
  \vspace{10pt}
  \begin{tabular}{lccc}
    \toprule
    object \textbackslash{} action & \textit{stay} & \textit{eat} & \textit{run} \\
    \midrule
    \textit{none} & $(0.3,0.35,0.35a,0.35b)$ & $(0.3,0.35,0.35a,0.35b)$ & $(0.3,0.25,0.45a,0.45b)$ \\
    \textit{predator} & $(0.45,0.2,0.35a,0.35b)$ & $(0.45,0.2,0.35a,0.35b)$ & $(0.55,0.1,0.35a,0.35b)$ \\
    \textit{prey} & $(0.25,0.35,0.4a,0.4b)$ & $(0.45,0.35,0.2a,0.2b)$ & $(0.45,0.35,0.2a,0.2b)$ \\
    \textit{rotten food} & $(0.25,0.35,0.4a,0.4b)$ & $(0.45,0.35,0.2a,0.2b)$ & $(0.45,0.35,0.2a,0.2b)$ \\
    \bottomrule
  \end{tabular}
  \vspace{5pt}
  \label{table_objtran}
  \vspace{-7pt}
\end{table}

\subsection{Hidden variable (\textit{hunger}, \textit{sickness}) transition table}

\begin{table}[H]
\vspace{-13pt}
  \caption{Hidden variable transition table. 2-tuple in each cell is the change of \textit{hunger} and \textit{sickness}
  for each state-action pair with respect to the previous values. R denotes reset to $0$.}
  \centering
  \vspace{10pt}
  \begin{tabular}{lccc}
    \toprule
    object \textbackslash{} action & \textit{stay} & \textit{eat} & \textit{run} \\
    \midrule
    \textit{none} & $(0.05,-0.05)$ & $(0.05,0.05)$ & $(0.15,0.1)$ \\
    \textit{predator} & $(0.05,-0.05)$ & $(\text{R},0.05)$ & $(0.15,0.1)$ \\
    \textit{prey} & $(0.05,-0.05)$ & $(\text{R},0.05)$ & $(0.15,0.1)$ \\
    \textit{rotten food} & $(0.05,-0.05)$ & $(\text{R},0.25)$ & $(0.15,0.1)$ \\
    \bottomrule
  \end{tabular}
  \vspace{5pt}
  \label{table_hiddentran}
  \vspace{-7pt}
\end{table}

\subsection{Terminal condition}

\begin{table}[H]
  \caption{Terminal probability table.}
  \centering
  \vspace{10pt}
  \begin{tabular}{lcccccccccc}
    \toprule
    \textbf{object} \textbackslash{} \textbf{action} & \textit{stay} & \textit{eat} & \textbf{hidden} \textbackslash{} \textbf{value} & $\le 0.7$ & $0.75$ & $0.8$ & $0.85$ & $0.9$ & $0.95$ & $1.0$ \\
    \midrule
    \textit{predator} & $0.6$ & $0.7$ & \textit{hunger} & $0.0$ & $0.05$ & $0.1$ & $0.2$ & $0.4$ & $0.7$ & $1.0$ \\
     &  &  & \textit{sickness} & $0.0$ & $0.05$ & $0.1$ & $0.2$ & $0.4$ & $0.7$ & $1.0$ \\
    \bottomrule
  \end{tabular}
  \vspace{5pt}
  \label{table_termination}
  \vspace{-7pt}
\end{table}

\section{Network architecture details}
\label{networkdetail}
Our network consists of three parts: CNN, RNN, and MLP. The CNN has three convolution layers with kernel sizes $(6, 4, 3)$, number of channels $(16, 32, 32)$, and strides $(3, 2, 1)$ without zero-padding. ReLU activation is used for convolution layers.
It is followed by a linear layer with LeakyReLU activation
that yields a 4-dimensional output.
The RNN consists of two hidden linear layers with 64 hidden neurons. We use ReLU for hidden activation, and Sigmoid for output activation. The MLP consists of one hidden linear layer with 64 hidden neurons. ReLU is used as the activation function.  

\newpage

\section{Algorithm}
\label{appen_algorithm}

\begin{algorithm}[h]
\caption{Update method (sequential update)}
\begin{algorithmic}[1]
\State Initialize episodic replay memory $D$ with size $|D|$
\State Initialize the mini-batch $B$ with size $|B|$, sub-episode length $L$
\State Initialize the online and target MLP parameters $\theta$ and $\theta'=\theta$.

\State Initialize the online and target RNN network parameters $w$ and $w'=w$.

\State Initialize the online and target CNN network parameters $\eta$ and $\eta'=\eta$.

\While{not done}
  \For{$l =0,1,\ldots,n-1$}
    \State Collect episode 
    \[e=((v_{1},h_{1},a_{1},r_{1}),(v_{2},\hat h_{2},a_{2},r_{2}),\ldots ,(v_T,\hat h_T,a_T,r_T))\]
    \qquad \quad with $\epsilon$-greedy policy, where $(\hat h_2 ,\hat h_3 , \ldots ,\hat h_T )$ is generated using 
    \[
    \hat{h}_{t}=\text{RNN}_w(\hat{h}_{t-1},\text{CNN}_\eta(v_{t-1}),a_{t-1}).
    \]
    \If {$\text{length}(e) > L$} 
      \State Split the episode $e$ into sub-episodes of length $L$
      \[
      \qquad \qquad \tilde e_1 = ( g_{1}, g_{2}, \ldots , g_{L}), \ldots, \tilde e_{i+1} = ( g_{iL+1}, g_{iL+2}, \ldots , g_{(i+1)L}) , \ldots, \tilde e_{last} = (  \ldots , g_{T}) 
      \] 
      \qquad \qquad \; where $g_i =
      \begin{cases}
      (v_1, h_1, a_1, r_1), \textrm{  if } i=1 \\
      (v_i, \hat{h}_i, a_i, r_i), \textrm{  otherwise}
      \end{cases}
      $\\
      \qquad \qquad \; and store the sub-episodes in $D$
    \Else 
      \State Store episode $e$ in $D$ 
    \EndIf
     
  \EndFor
  
  \For{$i=0,1,\ldots,m-1$}
    \State Sample mini batch $B$ from from $D$
    \For{$t=0,1,\ldots,L-1$}
      \State Perform a BPTT gradient update on empirical Bellman loss with respect to $\theta,w,\eta$
\[
\qquad \qquad \; \mathcal{L}(\theta,w,\eta,t):=\frac{1}{2}\frac{1}{|B|}\sum_{e\in B} {(r_t+\gamma\max_a Q_{\theta',\eta',w'} (v_{t+1},h_1,a)-Q_{\theta,\eta,w} (v_t,h_1,a_t))} 
\]
\qquad \qquad \; where $e$ is the episode of length $L$
\[
e = ((v_1,h_1,a_1,r_1),\ldots,(v_L,h_L,a_L,r_L)) \in B
\]
\[
Q_{\theta,\eta,w}(v_t, h_1, a_t) = {\rm MLP}_{\theta}({\rm CNN}_{\eta}(v_t),H_{w,\eta,t}(e),a_t)
\]
\[
H_{w,\eta,t}(e):= G_{w,\eta,v_{t-1},a_{t-1}}\circ\cdots \circ G_{w,\eta,v_1,a_1} (h_1 )
\]
\qquad \qquad \; and 
\[
G_{w,\eta,v_{t-1},a_{t-1}}(\cdot):={\rm RNN}_w(\cdot,{\rm CNN}_\eta(v_{t-1}),a_{t-1})
\]

    \EndFor
  \EndFor
  \State Update target $\theta',w',\eta'$ from $\theta,w,\eta$ with soft update
\EndWhile

\end{algorithmic}
\end{algorithm}

\newpage

\section{Vision-Action dependencies}
\label{va-decouple}

\begin{figure}[H]
\includegraphics[width=0.61\textheight]{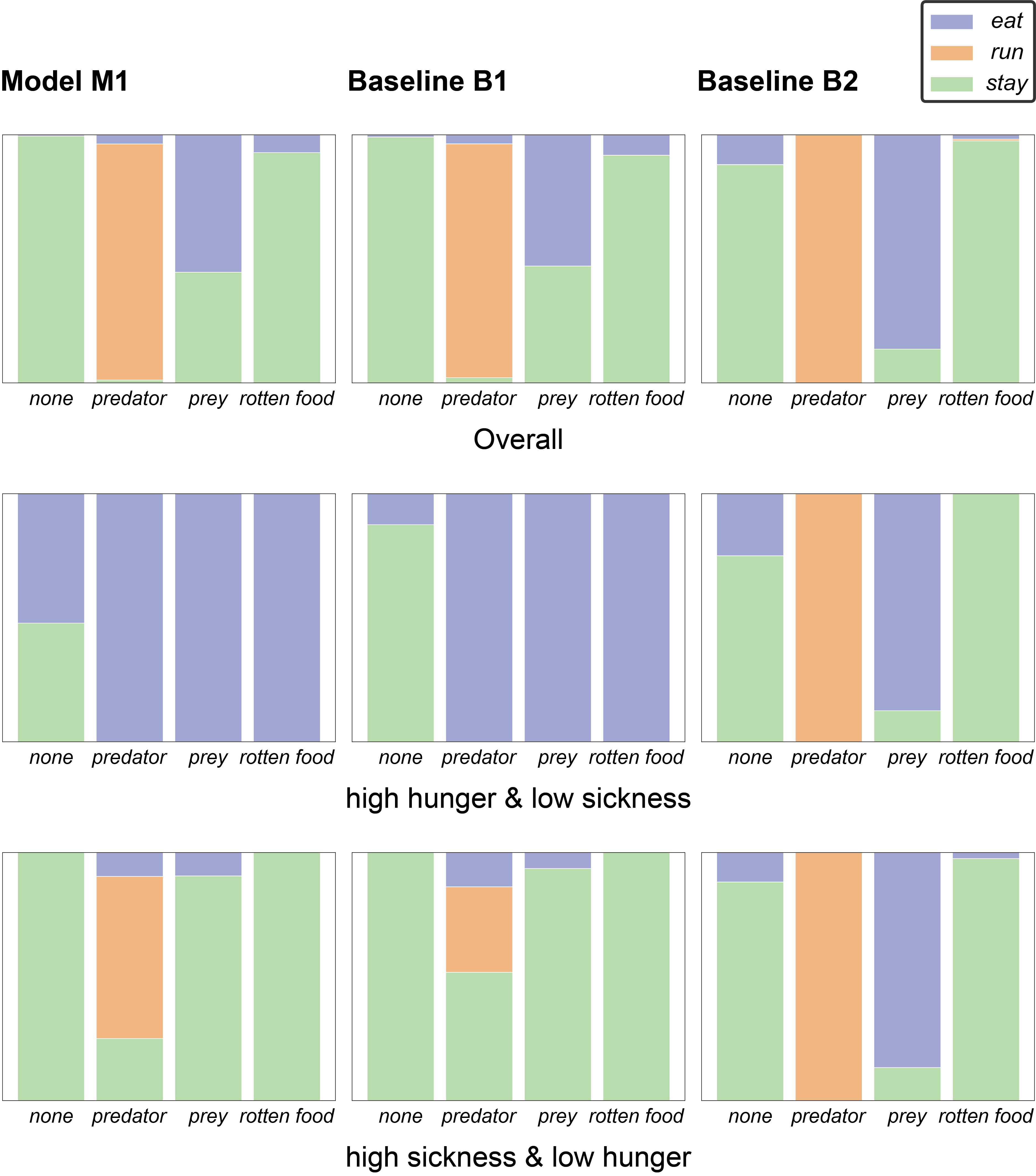}
\caption{vision-action dependencies of M1 (permutation 1), B1 and B2 after training. 
}
\label{fig_vision_action_decouple}
\end{figure}

Figure~\ref{fig_vision_action_decouple} shows that the agent chooses different actions for the same object class depending on the \textit{hunger} and \textit{sickness} level (i.e., vision input and the agent's action are decoupled).
The \textit{hunger} level was categorized as high if above 0.8 and as low if below 0.4. The \textit{sickness} level was categorized as high if above 0.7 and as low if below 0.65. 
In M1 and B1, the probability to choose each action depends largely on the \textit{sickness} when \textit{hunger} values, whereas it remains nearly unchanged in B2. This shows that the vision input and the optimal action are decoupled in the environment, which indicates that learning to choose the optimal action 
is a different task than learning to classify the images.

\newpage

\end{document}